\documentclass[twoside,11pt]{article}

\usepackage[preprint]{jmlr2e}
\usepackage{acro}
\usepackage{hyperref}
\usepackage{booktabs}
\usepackage[symbol]{footmisc}
\usepackage[utf8]{inputenc}

\usepackage{pdfrender}
\newcommand*{\boldcheckmark}{%
  \textpdfrender{
    TextRenderingMode=FillStroke,
    LineWidth=.5pt, 
  }{\checkmark}%
}

\usepackage{amssymb}
\usepackage{pifont}%

\newcommand{\xmark}{-}%

\jmlrheading{1}{2000}{1-48}{4/00}{10/00}{meila00a}{The Authors}

\ShortHeadings{PyKEEN 1.0}{Ali, Berrendorf, Hoyt, and Vermue}
\firstpageno{1}
\DeclareAcronym{aucroc}{
    short = AUC-ROC ,
    long = area under the Receiver Operating Characteristic curve
}
\DeclareAcronym{aucpr}{
    short = AUC-PR ,
    long = area under the precision-recall curve
}
\DeclareAcronym{amr}{
    short = AMR ,
    long = adjusted mean rank
}
\DeclareAcronym{amrr}{
    short = AMRR ,
    long = adjusted mean reciprocal rank
}
\DeclareAcronym{bns}{
    short = BNS ,
    long = Bernoulli negative sampling
}
\DeclareAcronym{cp}{
    short = CP ,
    long = canonical polyadic ,
}
\DeclareAcronym{cnn}{
    short = CNN,
    long = convolutional neural network,
}
\DeclareAcronym{cwa}{
	short = CWA ,
	long  = closed world assumption ,
}
\DeclareAcronym{gcn}{
	short = GCN ,
	long  = graph convolutional network ,
}
\DeclareAcronym{hpo}{
	short = HPO ,
	long  = hyper-parameter optimization ,
}
\DeclareAcronym{kg}{
	short = KG ,
	long  = knowledge graph ,
}
\DeclareAcronym{kge}{
	short = KGE ,
	long  = knowledge graph embedding ,
}
\DeclareAcronym{kgem}{
	short = KGEM ,
	long  = knowledge graph embedding model ,
}
\DeclareAcronym{lcwa}{
	short = LCWA ,
	long  = local closed world assumption ,
}
\DeclareAcronym{mr}{
    short = MR ,
    long = mean rank
}
\DeclareAcronym{fmr}{
    short = FMR ,
    long = filtered mean rank
}
\DeclareAcronym{mrr}{
    short = MRR ,
    long = mean reciprocal rank
}
\DeclareAcronym{owa}{
	short = OWA ,
	long  = open world assumption ,
}
\DeclareAcronym{rgcn}{
    short = RGC-N ,
    long = Relational Graph Convolutional Network
}
\DeclareAcronym{se}{
    short = SE ,
    long = Structured Embedding
}
\DeclareAcronym{smbo}{
	short = SMBO ,
	long  = Sequential Model-Based Global Optimization
}
\DeclareAcronym{smm}{
	short = SMM ,
	long  = semantic matching model ,
}
\DeclareAcronym{tdm}{
	short = TDM ,
	long  = translational distance model ,
}
\DeclareAcronym{tpe}{
	short = TPE ,
	long  = tree-structured parzen estimator ,
}
\DeclareAcronym{um}{
	short = UM ,
	long  = Unstructured Model ,
	cite = Glorot2013 ,
}
\DeclareAcronym{umls}{
	short = UMLS ,
	long  = Unified Medical Language System ,
	cite = mccray2003upper
}
\DeclareAcronym{yago}{
	short = YAGO ,
	long  = Yet Another Great Ontology ,
	cite = rebele2016yago
}

\DeclareAcronym{cel}{
	short = CEL ,
	long  = cross entropy loss ,
}

\title{PyKEEN 1.0: A Python Library for Training and Evaluating Knowledge Graph Embeddings}

\author{\name Mehdi Ali\footnotemark[1] \email mehdi.ali@cs.uni-bonn.de \\
    \addr Smart Data Analytics Group, University of Bonn \& Fraunhofer IAIS
    \AND
    \name Max Berrendorf\footnotemark[1] \email berrendorf@dbs.ifi.lmu.de\\
    \addr Ludwig-Maximilians-Universit\"at M\"unchen
    \AND
    \name Charles Tapley Hoyt\footnotemark[1] \email charles.hoyt@envedatx.com \\
    \addr Enveda Therapeutics
    \AND
    \name Laurent Vermue\footnotemark[1] \email lauve@dtu.dk \\
    \addr Technical University of Denmark
    \AND
    \name Sahand Sharifzadeh \email{sharifzadeh@dbs.ifi.lmu.de}\\
    \addr Ludwig-Maximilians-Universit\"at M\"unchen
    \AND
    \name Volker Tresp \email{volker.tresp@siemens.com}\\
    \addr Ludwig-Maximilians-Universit\"at M\"unchen \& Siemens AG
    \AND
    \name Jens Lehmann \email jens.lehmann@cs.uni-bonn.de \\
    \addr Smart Data Analytics Group, University of Bonn \& Fraunhofer IAIS\\
}

\begin{document}
\maketitle

\begin{abstract}

Recently, knowledge graph embeddings (KGEs) received significant attention, and several software libraries have been developed for training and evaluating KGEs.
While each of them addresses specific needs, we re-designed and re-implemented PyKEEN, one of the first KGE libraries, in a community effort. 
PyKEEN 1.0 enables users to compose knowledge graph embedding models based on a wide range of interaction models, training approaches, loss functions, and permits the explicit modeling of inverse relations.
Besides, an automatic memory optimization has been realized in order to exploit the provided hardware optimally, and through the integration of Optuna, extensive hyper-parameter optimization functionalities are provided.
\footnotetext[1]{Equal contribution.}

\end{abstract}


\section{Introduction}

Knowledge graphs (KGs) encode knowledge as a set of triples $\mathcal{K} \subseteq \mathcal{E} \times \mathcal{R} \times \mathcal{E}$ where $\mathcal{E}$ denotes the set of entities and $\mathcal{R}$ the set of relations.
Knowledge graph embedding models (KGEMs) learn representations for entities and relations of KGs in vector spaces while preserving the graph's structure.
The learned embeddings can support machine learning tasks such as entity clustering, link prediction, entity disambiguation as well as downstream tasks such as question answering and recommendation~\citep{nickel2015review, wang2017knowledge}.

Most publications of KGEMs are accompanied by reference implementations, but they often lack the finesse required for general usability.
Existing software packages that provide implementations for different KGEMs usually lack entire composability: model architectures (or interaction models), training approaches, loss functions, and the usage of explicit inverse relation cannot arbitrarily be combined.
The full composability of KGEMs is fundamental for assessing the performance of KGEMs because it allows assessing single components individually on the model's performance instead of attributing a performance increase solely to the model architecture, which is misleading~\citep{ruffinelli2020you,ali2020benchmarking}.
Besides, often only limited functionalities are provided, e.g., a small number of KGEMs are supported, or functionalities such as HPO are missing.
For instance, in PyKEEN~\citep{ali2019biokeen,ali2019keen} one of the first software packages for KGEMs, models can only be trained under the stochastic local closed-world assumption, the evaluation procedure was too slow for larger KGs, and it was designed to be mainly used through a command-line interface rather than programmatically in order to facilitate its usage for non-experts.
This motivated the development of a reusable software package comprising several KGEMs and related methodologies that is entirely configurable.

Here, we present PyKEEN (Python KnowlEdge EmbeddiNgs) 1.0, a community effort in which PyKEEN has been re-designed and re-implemented from scratch to overcome the mentioned limitations, make models entirely configurable, and to extend it with more interaction models and other components.

\section{System Description}

In PyKEEN 1.0, a KGEM is considered as a composition of four components that can flexibly be combined: an interaction model (or model architecture), a loss function, a training approach, and the usage of inverse relations.
PyKEEN 1.0 currently supports 23 interaction models, seven loss functions, four regularizers, two training approaches, HPO, six evaluation metrics, and 13 built-in benchmarking datasets.
It can readily import additional datasets that have been pre-stratified into train/test/evaluation and generate appropriate splits for unstratified datasets.
Additionally, we implemented an automatic memory optimization that ensures that the available memory is best utilized.

\paragraph{\textbf{Composable KGEMs}}

To ensure the composability of KGEMs, the interaction models, loss functions, and training approaches are separated from each other and implemented as independent submodules, whereas the modeling of inverse relations is handled by the interaction models.
Our modules can be arbitrarily replaced because we ensured through inheritance that all interaction models, loss functions, and training approaches follow unified APIs, which are defined by \textit{pykeen.model.Model}, \textit{pykeen.loss.Loss}, and \textit{pykeen.training.TrainingLoop}. 
Currently, we provide implementations of 23 interaction models, the most common loss functions used for training KGEMs including the \textit{binary-cross entropy}, \textit{cross entropy}, \textit{mean square error}, \textit{negative-sampling self-adversarial loss}, and the \textit{softplus loss}, as well as the \textit{local closed-world assumption} and the \textit{stochastic local closed-world assumption} training approach~\citep{nickel2015review}.
It is known that some interaction models (e.g., ConvE) benefit from being explicitly trained with inverse relations, i.e., for each relation $r \in \mathcal{R}$ an inverse relation $r_{inv}$ is introduced, and the task of predicting the head entity of a $(r,t)$-pair becomes the task of predicting the tail entity of the corresponding inverse pair $(t, r_{inv})$.
Therefore, in PyKEEN 1.0, we enable users to train each interaction model with explicit inverse relations.

\paragraph{\textbf{Evaluation}}

KGEMs are usually evaluated on the task of link prediction.
Given $(h, r)$ (or $(r, t)$), all possible entities $\mathcal{E}$ are considered as tail (or head) and ranked according to the KGEMs interaction model.
The individual ranks are commonly aggregated to mean rank, mean reciprocal rank, and hits@k.
However, these metrics have been realized differently throughout the literature based on different definitions of the rank~\citep{berrendorf2020interpretable}, leading to difficulties in reproducibility and comparability~\citep{akrami2018re}.
The three most common rank definitions are the \textit{average rank}, \textit{optimistic rank}, and \textit{pessimistic rank}.
In PyKEEN 1.0, we explicitly compute the aggregation metrics for all common rank definitions, \emph{average}, \emph{optimistic}, and \emph{pessimistic}, allowing inspection of differences between them.
This can help to reveal cases where the model predicts exactly equal scores for many different triples, which is usually an undesired behavior.
In addition, we support the recently proposed \emph{adjusted mean rank}~\citep{berrendorf2020interpretable}, which allows comparing results across differently sized datasets, as well as offering an interface to use all metrics implemented in scikit-learn~\citep{scikit-learn}, including AUC-PR and AUC-ROC.

\paragraph{\textbf{Hyper-Parameter Optimization}}

We integrated Optuna~\citep{akiba2019optuna} as the hyper-parameter optimization (HPO) framework to enable PyKEEN 1.0 to take advantage of its wide range of HPO functionalities (i.e., grid search, random search, tree-parzen estimator).
To optimize the hyper-parameters on the validation set, we implemented early stopping.
Besides, we implemented an HPO workflow that enables users to effectively find an appropriate set of hyper-parameters on the validation set, and train and evaluate the final model on the test set $n$ times to measure the robustness of the model.

\paragraph{\textbf{Automatic Memory Optimization}}

Allowing high computational throughput while ensuring that the available hardware memory is not exceeded during training and evaluation requires the knowledge of the maximum possible training and evaluation batch size for the current model configuration.
However, determining the training and evaluation batch sizes is a tedious process, and not feasible when a large set of heterogeneous experiments are run.
Therefore, we implemented an automatic memory optimization step that computes the maximum possible training and evaluation batch sizes for the current model configuration and available hardware before the actual experiment starts.
If the user-provided batch size is too large for the used hardware, the automatic memory optimization determines the maximum sub-batch size for the training.



\paragraph{\textbf{Extensibility}}

Because we defined a uniform API for each interaction model, any new model can be integrated by following the API of the existing models (\textit{pykeen.models}). 
Similarly, the remaining components, e.g., regularizers, and negative samplers follow a unified API, so that new modules can be smoothly integrated.

\paragraph{\textbf{Community Standards}}

PyKEEN 1.0 relies on several community-oriented tools to ensure it is accessible, reusable, reproducible, and maintainable.
It is implemented for Python 3.7+ using the PyTorch package.
It comes with a suite of thorough unit tests that are automated with \href{https://docs.pytest.org/en/latest/}{PyTest}, \href{https://tox.readthedocs.io/en/latest/}{Tox}, run in a continuous integration setting on Travis-CI, and are tracked over time using \url{codecov.io}.
Code quality is ensured with flake8 and careful application of the \href{https://guides.github.com/introduction/flow/}{GitHub Flow} development workflow.
Documentation is quality checked by \href{https://launchpad.net/doc8}{doc8}, built with \href{https://www.sphinx-doc.org}{Sphinx}, and hosted on \url{ReadTheDocs.org}.

\section{Comparison to Related Software}

Table~\ref{tab:comparisson} depicts the most popular KGE frameworks and their features.
It shows that PyKEEN 1.0 compared to related software packages emphasize on both, full composability of KGEMs and extensive functionalities, i.e., a large number of supported interaction models, and extensive evaluation and HPO functionalities.
Finally, PyKEEN 1.0 is the only library that performs an automatic memory optimization that ensures that the memory is not exceeded during training and evaluation.
GraphVite, DGL-KE, and PyTorch-BibGraph focus on scalability, i.e., they provide support for multi-GPU/CPU or/and distributed training, but focus less on compositionality and extensibility. For instance, PyTorch-BigGraph supports only a small number of interaction models that follow specific computation blocks.

\begin{table}
\centering
\tiny
\caption{An overview of the functionalities of PyKEEN 1.0 and similar libraries.
\textbf{ES} refers to early stopping, \textbf{TA} to training approach, \textbf{Inv. Rels.} to the explicit modeling of inverse relations, \textbf{AMO} to automatic memory optimization, \textbf{MGS} to multi-GPU support, and \textbf{DTR} to distributed training.} \label{tab:comparisson}
\begin{tabular}{lclclcccccc}
\toprule
Library    & \begin{tabular}[c]{@{}l@{}} Models\end{tabular} & HPO                                               & \begin{tabular}[c]{@{}l@{}}ES\end{tabular} & \begin{tabular}[c]{@{}l@{}}Evaluation \\ Metrics\end{tabular}                               & \begin{tabular}[c]{@{}l@{}}Set\\ TA\end{tabular} & \begin{tabular}[c]{@{}l@{}}Inv. \\ Rels.\\\end{tabular} & \begin{tabular}[c]{@{}l@{}}Set\\ Loss\\ Fct.\end{tabular} & \begin{tabular}[c]{@{}l@{}}AMO\end{tabular}  &
MGS & DTR \\
\midrule
\begin{tabular}[c]{@{}l@{}}AmpliGraph\\\citep{ampligraph}\end{tabular} 
 & 6                                                    & GS                                                & \checkmark                                                       & \begin{tabular}[c]{@{}l@{}}MR, MRR,\\ Hits@k\end{tabular}                                & \xmark                                                                              & \checkmark                                                          & \checkmark                                                                 & \xmark & \xmark & \xmark                                                                   \\
\begin{tabular}[c]{@{}l@{}}DGL-KE\\\citep{zheng2020dgl}\end{tabular} 
& 6 & \xmark & \xmark & \begin{tabular}[c]{@{}l@{}}MR, MRR \\ Hits@k\end{tabular}  & \xmark & \xmark & \checkmark & \xmark  & \checkmark & \checkmark \\
\begin{tabular}[c]{@{}l@{}}GraphVite\\\citep{zhu2019graphvite}\end{tabular} 
& 6  & \xmark   & \xmark  & \begin{tabular}[c]{@{}l}MR, MRR,\\ Hits@k,\\ AUC-ROC\end{tabular} & \xmark  & \xmark  & \xmark   & \xmark  & \checkmark & \xmark\\
\begin{tabular}[c]{@{}l@{}}LibKGE\\\citep{ruffinelli2020you}\end{tabular}
 & 10                                                    & \begin{tabular}[c]{@{}l@{}}GS, RS, TPE\end{tabular}                                               & \checkmark                                                       & \begin{tabular}[c]{@{}l@{}}MR, MRR,\\ Hits@k\end{tabular}                                & \checkmark                                                                               & \checkmark                                                          & \checkmark                                                                 & \xmark                                                                   & \xmark & \xmark \\ 
\begin{tabular}[c]{@{}l@{}}OpenKE\\\citep{han2018openke}\end{tabular}
& 11                                                   & \xmark                                                & \xmark                                                        & \begin{tabular}[c]{@{}l@{}}MR, MRR,\\ Hits@k\end{tabular}                                & \xmark                                                                              & \xmark                                                               & \checkmark                                                                 & \xmark                                                                    & \xmark & \xmark \\ 
\begin{tabular}[c]{@{}l@{}}PyTorch-BigGraph\\\citep{pbg}\end{tabular}
 & 4  & \xmark  & \xmark  & \begin{tabular}[c]{@{}l}MR, MRR,\\ Hits@k, \\AUC-ROC\end{tabular}  & \xmark   & \xmark   & \checkmark & \xmark  & \checkmark & \checkmark\\
\begin{tabular}[c]{@{}l@{}}Pykg2vec\\\citep{yu2019pykg2vec}\end{tabular}& 18                                                   & TPE                                               & \xmark                                                        & \begin{tabular}[c]{@{}l@{}}MR, Hits@k\end{tabular}                                       & \xmark                                                                              & \xmark                                                               & \xmark                                                                  & \xmark                                                                    & \xmark & \xmark\\ 
\begin{tabular}[c]{@{}l@{}}PyKEEN\\\citep{ali2019keen}\end{tabular}& 10                                                   & \begin{tabular}[c]{@{}l@{}}RS\end{tabular} & \xmark                                                       & \begin{tabular}[c]{@{}l@{}}MR, Hits@k\end{tabular} & \xmark                                                                             & \xmark                                                             & \xmark                                                                 & \xmark                                                                   & \xmark & \xmark\\ 
\textbf{PyKEEN 1.0} & \textbf{23}                                                   & \begin{tabular}[c]{@{}l@{}}\textbf{GS, RS, TPE}\end{tabular} & \boldcheckmark                                                       & \begin{tabular}[c]{@{}l@{}}\textbf{MR, MRR}\\\textbf{AMR},\\\textbf{Hits@k}, \\ \textbf{AUC-PR},\\\textbf{AUC-ROC}\end{tabular} & \boldcheckmark                                                                             & \boldcheckmark                                                             & \boldcheckmark                                                                 & \boldcheckmark                                                                   & \xmark & \xmark\\
\bottomrule
\end{tabular}
\end{table}

\section{Availability and Maintenance}

PyKEEN 1.0 is publicly available under the MIT License at \url{https://github.com/pykeen/pykeen}, and is distributed through the \href{http://pypi.org/project/pykeen}{Python Package Index}.
It will be maintained by the core developer team that is supported by the Smart Data Analytics research group (University of Bonn), Fraunhofer IAIS, Enveda Therapeutics, Munich Center for Machine Learning (MCML), Siemens, and the Technical University of Denmark (section for Cognitive Systems and section for Statistics and Data Analysis).
The project is funded by the German Federal Ministry of Education and Research (BMBF) under Grant No. 01IS18036A and Grant No. 01IS18050D (project MLWin) as well as the Innovation Fund Denmark with the Danish Center for Big Data Analytics driven Innovation (DABAI) which ensures the maintenance of the project in the next years.


%



\vskip 0.2in
\bibliography{lib}


\end{document}